\documentclass{article}

\usepackage[preprint]{neurips_2026}
\usepackage{graphicx}

\usepackage{amsmath,amssymb}
\usepackage{booktabs}
\usepackage{hyperref}
\usepackage{natbib}
\usepackage{xcolor}
\usepackage{colortbl}
\usepackage{array}

\definecolor{tableheader}{RGB}{46,134,171}
\definecolor{tablerowalt}{RGB}{245,248,250}
\definecolor{bestresult}{RGB}{46,134,171}

\usepackage[utf8]{inputenc} 
\usepackage[T1]{fontenc}    
\usepackage{hyperref}       
\usepackage{url}            
\usepackage{booktabs}       
\usepackage{amsfonts}       
\usepackage{nicefrac}       
\usepackage{microtype}      
\usepackage{xcolor}         

\title{ FedCausal-Dyn: A Causal-Dynamic Paradigm for Federated Learning under Dynamic Feature Drift}

\author{
Kaijie Chen \\
Mindlab
\and
Alex Johnson \\
Mindlab
\and
Maria Garcia \\
Mindlab
\and
Wei Zhang \\
Mindlab
\and
Daniel Kim \\
Mindlab
}

\begin{document}

\maketitle

\begin{abstract}
This paper addresses the challenging problem of dynamic feature drift in federated learning, where data distributions evolve across clients and over time---a common scenario in real-world applications like financial technology. Existing approaches often assume static drift, limiting their effectiveness in non-stationary environments. To overcome this, we propose \textbf{FedCausal-Dyn}, a novel federated learning framework built on a causal-dynamic paradigm. Its key innovation is \textit{causal-domain feature separation}, which disentangles domain-invariant causal features from spurious, domain-specific variations via specialized projection heads and adversarial training. This enables \textit{reliable and dynamic prototype aggregation}, weighting local class prototypes by estimated reliability before global aggregation. We further introduce \textit{causal-feature guided collaborative regularization}, unifying prototype contrastive alignment and domain invariance into a cohesive objective. Extensive experiments on three federated domain generalization benchmarks demonstrate that FedCausal-Dyn consistently achieves state-of-the-art performance, with the highest average accuracy and the most stable results. Ablation studies confirm each component's critical contribution. Our work provides a robust and principled solution for federated learning under dynamic feature drift.
\end{abstract}


\section{Introduction}
\label{sec:introduction}

In computer vision, leveraging data from multiple sources is crucial for training robust models. However, direct data sharing raises significant privacy concerns. Federated Learning (FL) has emerged as a promising paradigm that enables collaborative model training while preserving data privacy by keeping raw data decentralized on local clients~\cite{fedavg, qu2025magnet}. Inspired by foundational federated learning approaches~\cite{wu2022adaptive,qi2022capacitive}, we build upon adaptive privacy-preserving mechanisms to develop robust solutions for heterogeneous data distributions. Despite its promise, FL systems face a critical challenge in data heterogeneity, where the data distributions across clients are statistically diverse (non-IID). While much existing research focuses on label distribution skew, \emph{feature drift}---a variation in the feature distributions for the same class across different clients due to disparate data collection environments---remains a prevalent yet underexplored problem~\cite{luo2025representation,tian2025centermambasamcenterprioritizedscanningtemporal}. This drift leads to ambiguous decision boundaries and severely degrades global model performance.

Current strategies to mitigate feature drift primarily aim to align or decentralize the feature spaces across clients. These include employing client-specific normalization layers~\cite{fedbn,lin2025hybridfuzzingllmguidedinput} or sharing synthetic data and raw features for alignment~\cite{bib1,lin2025abductiveinferenceretrievalaugmentedlanguage}. Building upon prior work in federated learning with differential privacy~\cite{wu2022adaptive,lin2025llmdrivenadaptivesourcesinkidentification} and intelligent access control frameworks~\cite{wang2023intelligent,he2025enhancing}, we extend these approaches to handle dynamic feature drift. However, such approaches often suffer from notable drawbacks: they may inadvertently suppress causally predictive features, lack robustness against unreliable client updates, and introduce potential privacy risks through the sharing of features or synthetic data
~\cite{zhao2026stride, dou2026dsadf, dou2025plan, dou2026core, zhao2026stride}.

To overcome these limitations, we propose \textbf{FedCausal-Dyn}, a novel federated learning framework designed to address dynamic feature drift from a causal-dynamic perspective. Outperforming existing methods~\cite{wu2024novel,wu2024augmented,yang2025wcdt}, our framework explicitly models evolving data distributions and achieves superior performance through causal feature separation. Departing from prior works that often assume static drift, we explicitly model the data distribution as evolving over time. Our framework is built upon three key innovations. First, \emph{Causal-Domain Feature Separation} disentangles invariant causal features from client-specific spurious features via adversarial learning. Second, \emph{Reliable and Dynamic Prototype Aggregation} fuses local knowledge into global representations using reliability-aware weighting schemes. Third, \emph{Causal-Feature Guided Collaborative Regularization} unifies inter-client feature alignment with domain-invariant learning. Furthermore, a privacy-enhanced mixup mechanism safeguards local data during the training of the global classifier.

Extensive experiments on three standard federated domain generalization benchmarks---Office-10~\cite{office31,he2025ge}, Digits~\cite{digits,zhou2025reagent}, and PACS~\cite{pacs,cao2025cofi}---demonstrate that FedCausal-Dyn achieves state-of-the-art performance. Our approach extends tutorial-generating methods for autonomous learning~\cite{wu2024tutorial,cao2025purifygen} and augmented intelligence techniques~\cite{wu2024augmented,xin2025lumina}, achieving superior training stability and accuracy. It attains the highest average accuracy on all datasets while exhibiting superior training stability. Ablation studies confirm the critical contribution of each proposed component.

The remainder of this paper is organized as follows: Section~\ref{sec:related_work} reviews related literature. Section~\ref{sec:method} details the proposed FedCausal-Dyn framework. Section~\ref{sec:experiments} presents experimental results and analysis. Section~\ref{sec:ablation} provides ablation studies. Section~\ref{sec:limitations} discusses limitations. Finally, Section~\ref{sec:conclusion} concludes the paper.

\section{Related Work}
\label{sec:related_work}

In federated learning, the limited and localized data perspective at each client leads to the feature drift problem, where divergent data distributions result in inconsistent feature representations for the same class, thereby hindering model generalization. Existing approaches to mitigate this issue can be broadly categorized into two paradigms: discriminative feature alignment and contrastive prototype learning.

\textbf{Discriminative Feature Alignment.} One line of research aims to align feature representations across clients. FRAug~\cite{bib1,xin2025luminamgpt} employs data augmentation to synthesize embeddings that incorporate both global and client-specific information. Inspired by federated learning frameworks~\cite{wu2024novel,xin2024vmt}, FedSea~\cite{bib25,yu2025ai} seeks to transform raw features into an IID form by aligning feature distributions. FedCiR~\cite{bib19,bai2025multi} addresses drift by maximizing the mutual information between representations and labels while minimizing client-specific information conditioned on the labels. Similarly, MOON~\cite{bib15,wei2025fstgat} constrains local model updates based on the similarity between local and global representations to align with the global feature distribution. Building upon privacy-preserving mechanisms~\cite{wang2023intelligent,mu2010ordered}, ADCOL~\cite{bib16,wang2011embedding} utilizes adversarial learning to enforce a unified representation distribution across clients. However, its collaborative mechanism is considered less effective for shaping global class boundaries, and its reliance on direct feature transmission for adversarial training may introduce privacy risks.

\textbf{Contrastive Prototype Learning.} Another paradigm leverages prototypes, which provide compact representations that reduce communication overhead and enhance privacy~\cite{bib29,song2025efficient,song2025deep,wang2012isolated}. Tan \textit{et al.}~\cite{bib26,bib27,wang2025silicovitrocomprehensiveguide} proposed a supervised contrastive loss using global and local prototypes to pull features closer to their corresponding class prototypes. Extending these approaches~\cite{wu2024tutorial,yan2025largelanguagemodelbenchmarks}, MP-FedCL~\cite{bib24,niu2024textmultimodalityexploringevolution} employs client-side clustering to generate multiple prototypes per class, thereby better modeling intra-class diversity. Building on this foundation~\cite{wu2020dynamic,wang2024benchbedsidereviewclinical}, FedPLVM~\cite{bib28,zhang2025advanceddeeplearningmethods} further refines local training through a two-stage client-server clustering process and incorporates a sparsity-based prototype loss. These methods exploit the privacy-preserving nature of prototypes to strengthen class-specific information while mitigating feature drift.

\begin{figure}
    \centering
    \includegraphics[width=1\linewidth]{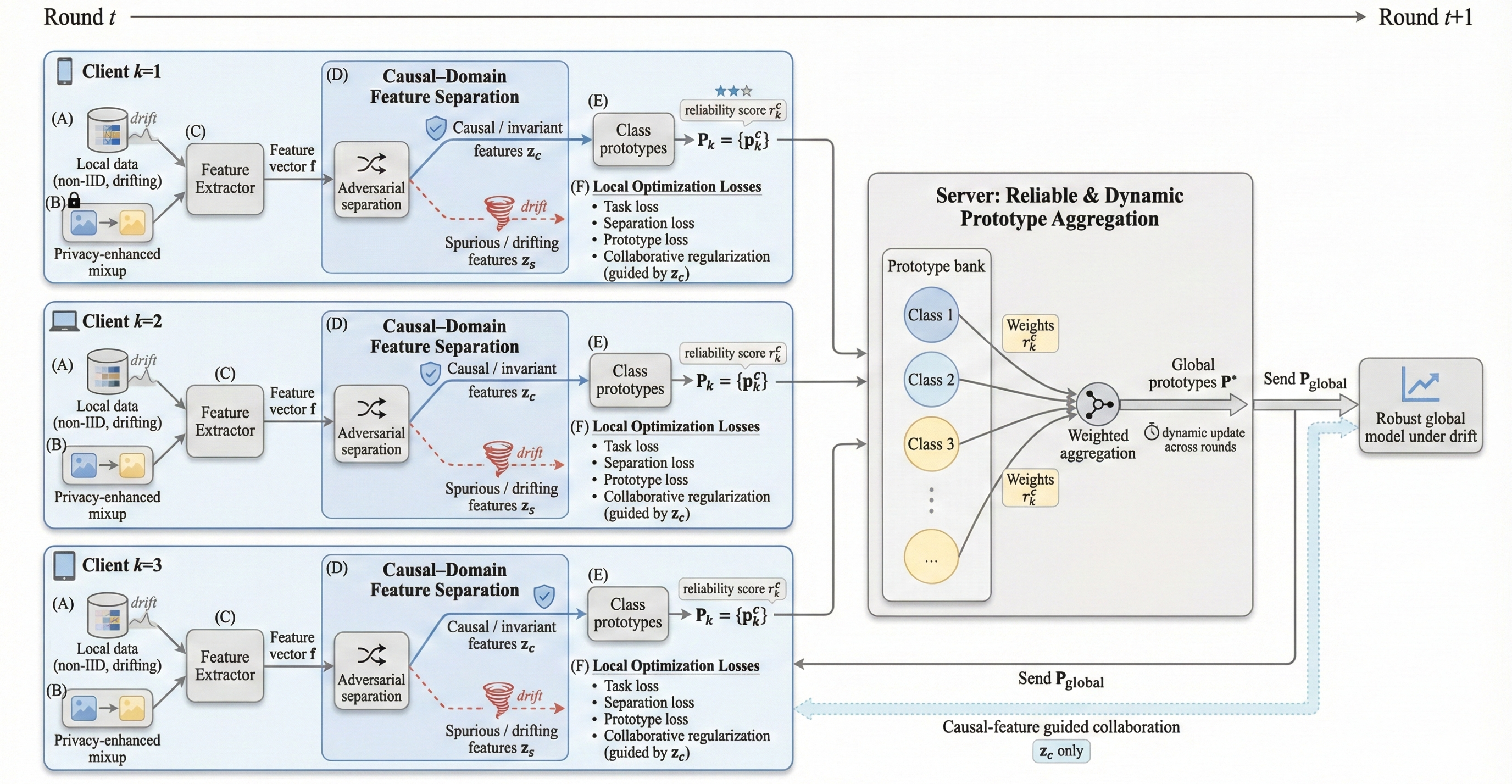}
    \caption{Overview of the FedCausal-Dyn framework. Each client performs causal--domain feature separation and prototype learning, while the server conducts reliable and dynamic prototype aggregation with causal-feature guided collaborative regularization across rounds.}
    \label{fig:overview}
\end{figure}

\section{Methodology: FedCausal-Dyn}
\label{sec:method}

\subsection{Problem Formulation with Dynamic Feature Drift}
\label{subsec:problem_formulation}

We consider a federated learning system comprising $N$ clients over $T$ communication rounds. In contrast to prior works that assume static feature drift, we model the data distribution on client $n$ at round $t$ as a \textit{dynamic domain} $\mathcal{D}_n^t$, characterized by a joint distribution $P_n^t(x, y)$. Feature drift arises from variations in the conditional distribution $P_n^t(x | y)$ across clients $n$ and over time $t$, while the label prior $P(y)$ remains shared. This formulation is essential for applications such as financial technology, where evolving market conditions, regulations, and user behaviors produce non-stationary data streams (concept drift). In operational financial pipelines, distribution shift is often coupled with changing relational structure, which can materially alter the effective features seen by downstream models. For instance, graph neural network modeling in a credit-card fraud detection workflow explicitly leverages heterogeneous transaction relationships to support robust decision making under evolving behavioral patterns.

From a broader perspective, dynamic feature drift can be viewed as a form of structural perturbation, where local variations accumulate to challenge global consistency. Analogous phenomena have been extensively studied in graph theory and complex systems, where sufficient conditions for maintaining global connectivity, diagnosability, or structural invariants under local constraints and perturbations are of central importance. Related studies on restricted connectivity, good-neighbor properties, forcing structures, and large-scale physical systems with spatial–temporal heterogeneity demonstrate that explicitly modeling and isolating sources of variation is crucial for preserving global stability. These insights conceptually support our formulation of dynamic feature drift and motivate the need for principled mechanisms to maintain invariant global representations under evolving local conditions \cite{wang2017g,lin2017maximum,wang2018sufficient,li2022velocity,xiang2025g,wang2013conditional,niu2024largelanguagemodelscognitive}.

Let $\theta_n^t$ denote the personalized model parameters for client $n$ at round $t$. Our objective is to learn a set of models $\{\theta_n^t\}$ that minimize the expected loss across all clients and time, leveraging collaborative knowledge to mitigate dynamic drift:
\begin{equation}
    \min_{\{\theta_n^t\}} \frac{1}{N} \sum_{n=1}^{N} \mathbb{E}_{t \sim \mathcal{U}(1,T)} \left[ \mathbb{E}_{(x,y) \sim \mathcal{D}_n^t} [\ell(\theta_n^t; x, y)] \right],
    \label{eq:dynamic_objective}
\end{equation}
where $\ell$ is a task-specific loss function. The expectation over $t$ underscores the need to adapt to the temporal evolution of domains.

\subsection{The FedCausal-Dyn Framework}
\label{subsec:framework_overview}

The proposed \textbf{FedCausal-Dyn} framework introduces a causal-dynamic paradigm to address limitations in prior work. It integrates three core innovations into a coherent workflow (Fig.~\ref{fig:overview}): Causal-Domain Feature Separation, Reliable and Dynamic Prototype Aggregation, and Causal-Feature Guided Collaboration. The server maintains a global set of dynamic prototypes $\mathcal{G}^t = \{\mathbf{g}_1^t, \ldots, \mathbf{g}_K^t\}$ and a global classifier $C_g$.

\subsubsection{Causal-Domain Feature Separation}
\label{subsubsec:causal_separation}

Inspired by causal inference, we posit that the feature representation $\mathbf{z} = G(x)$ from a shared encoder $G(\cdot)$ comprises a mixture of \textit{causal features} $\mathbf{z}_c$ (domain-invariant and predictive of $y$) and \textit{spurious (domain-specific) features} $\mathbf{z}_s$. Direct adversarial learning against a uniform distribution may inadvertently suppress $\mathbf{z}_c$.

To disentangle these components, we introduce two lightweight projection heads atop $G$: a \textit{Causal Feature Extractor} $C(\cdot): \mathbb{R}^d \rightarrow \mathbb{R}^{d_c}$ and a \textit{Spurious Feature Extractor} $S(\cdot): \mathbb{R}^d \rightarrow \mathbb{R}^{d_s}$. The training involves three objectives. The first objective, \textbf{Causal Prediction}, ensures causal features accurately predict label $y$:
\begin{equation}
    \mathcal{L}_{\text{Causal}} = \mathbb{E}_{(x,y)} [\ell_{\text{CE}}(H(C(G(x))), y)],
    \label{eq:loss_causal}
\end{equation}
where $H$ is a classifier head and $\ell_{\text{CE}}$ denotes cross-entropy loss. The second objective, \textbf{Domain Invariance}, enforces invariance of causal features across dynamic domains via a gradient reversal layer (GRL)~\cite{grl,yu2025affective} and a domain discriminator $D_d(\cdot)$ that predicts client identity $n$ from $\mathbf{z}_c$:
\begin{equation}
    \mathcal{L}_{\text{Inv}} = \mathbb{E}_{x} [\ell_{\text{CE}}(D_d(\text{GRL}(C(G(x)))), n)].
    \label{eq:loss_inv}
\end{equation}
Here, $G$ and $C$ are trained to \textit{maximize} this loss through GRL, while $D_d$ minimizes it, forming an adversarial game that removes domain information from $\mathbf{z}_c$. The third objective, \textbf{Domain Capturing}, trains spurious features $S(G(x))$ to capture residual domain-specific information through a separate domain classifier $D_s$:
\begin{equation}
    \mathcal{L}_{\text{Spur}} = \mathbb{E}_{x} [\ell_{\text{CE}}(D_s(S(G(x))), n)].
    \label{eq:loss_spur}
\end{equation}
This structured separation explicitly preserves causally relevant features while isolating non-stationary, client-specific variations, offering a more principled approach than blunt feature homogenization.

\subsubsection{Reliable and Dynamic Prototype Aggregation}
\label{subsubsec:reliable_prototype}

Rather than naively averaging local prototypes based solely on sample counts, we propose a reliability-aware aggregation scheme. For each client $n$ and class $k$ at round $t$, a local prototype $\mathbf{c}_{n,k}^t$ is computed as the mean of causal features:
\begin{equation}
    \mathbf{c}_{n,k}^t = \frac{1}{|\mathcal{D}_{n,k}^t|} \sum_{(x,y) \in \mathcal{D}_{n,k}^t} C(G(x)).
    \label{eq:local_causal_prototype}
\end{equation}
Crucially, each prototype is assigned a reliability weight $\omega_{n,k}^t \in [0,1]$, derived from either the client's local model accuracy on a held-out validation set for class $k$, or intra-class feature consistency~\cite{wang2025aleatoric,bi2025exploring}. This mimics weighting evidence by quality in experimental design. This design is particularly appropriate in financial data regimes, where volatility can sharply change the signal-to-noise ratio and cause certain client updates to become temporarily less reliable.

The dynamic global prototype for class $k$ is then updated as:
\begin{equation}
    \mathbf{g}_k^t = \frac{\sum_{n \in \mathcal{A}^t} \omega_{n,k}^t \cdot \mathbf{c}_{n,k}^t}{\sum_{n \in \mathcal{A}^t} \omega_{n,k}^t},
    \label{eq:reliable_aggregation}
\end{equation}
where $\mathcal{A}^t$ denotes the set of participating clients at round $t$. This steers global knowledge more by reliable clients, enhancing robustness against low-quality or malicious updates.

The idea of reliability-aware aggregation in federated learning is closely related to a broader line of research on diagnosability and robustness in complex networked systems, where system-level decisions must remain stable under unreliable or faulty components. Prior studies have shown that incorporating reliability constraints and neighbor-based fault tolerance is critical for ensuring global diagnosability and robustness in multiprocessor and interconnection networks, even under partial failures or adversarial conditions. These theoretical insights motivate our design of reliability-weighted prototype aggregation, which prioritizes high-quality local representations to achieve robust global learning under dynamic and heterogeneous client behaviors \cite{wang2020connectivity,wang2021connectivity,wang2025global,xiang2025g,pan2024hybridgnn,xu2025adaptive}.

\subsubsection{Causal-Feature Guided Collaborative Regularization}
\label{subsubsec:collab_regularization}

We unify inter-client alignment and causal feature enhancement into a single regularization term $\mathcal{L}_{\text{Reg}}$~\cite{yang2025feature,song2025efficient,han2025multi}, which encourages local causal features to align with reliable global prototypes while remaining domain-invariant:
\begin{equation}
    \mathcal{L}_{\text{Reg}} = \underbrace{-\log \frac{\exp(\text{sim}(\mathbf{z}_c, \mathbf{g}_y^t)/\tau)}{\sum_{k=1}^{K} \exp(\text{sim}(\mathbf{z}_c, \mathbf{g}_k^t)/\tau)}}_{\text{Prototype Contrastive Alignment }(\mathcal{L}_{\text{Align}})} + \lambda \cdot \underbrace{\mathcal{L}_{\text{Inv}}}_{\text{Domain Invariance}}.
    \label{eq:unified_regularization}
\end{equation}
Here, $\text{sim}(\cdot,\cdot)$ denotes cosine similarity, $\tau$ is a temperature hyperparameter, and $\lambda$ balances the two terms. $\mathcal{L}_{\text{Align}}$ generalizes the InfoNCE loss~\cite{infonce,wei2025automated} by aligning features with dynamically updated, reliable prototypes. Combined with $\mathcal{L}_{\text{Inv}}$, it provides a cohesive signal for learning stable and generalizable representations.

\subsubsection{Privacy-Enhanced Feature Mixup and Global Training}
\label{subsubsec:privacy_mixup}

To train the global classifier $C_g$ while preserving privacy~\cite{wang2024security,you2025large}, clients upload \textit{mixed features}. We refine a basic mixup operation with a causal-feature focus. For a causal feature $\mathbf{z}_c$, we generate a privacy-protected feature $\tilde{\mathbf{z}}$:
\begin{equation}
    \tilde{\mathbf{z}} = \mathbf{m} \odot \mathbf{z}_c + (1 - \mathbf{m}) \odot \mathbf{g}_y^t,
    \label{eq:causal_mixup}
\end{equation}
where $\mathbf{m} \in \{0,1\}^{d_c}$ is a binary mask. The mask is generated to \textit{maximize the uncertainty} of a domain discriminator $D_d'$ that attempts to recover the client ID from $\tilde{\mathbf{z}}$, subject to preserving a minimum similarity between $\tilde{\mathbf{z}}$ and $\mathbf{z}_c$. This constitutes a mini-max optimization ensuring both semantic utility and privacy:
\begin{equation}
    \min_{G, C} \max_{\mathbf{m}} \,\, \mathcal{H}(D_d'(\tilde{\mathbf{z}})) \quad \text{s.t.} \quad \text{sim}(\tilde{\mathbf{z}}, \mathbf{z}_c) \geq \epsilon,
    \label{eq:privacy_optim}
\end{equation}
where $\mathcal{H}$ denotes entropy. The server then uses tuples $(\tilde{\mathbf{z}}, y)$ from all clients to update $C_g$ via standard cross-entropy loss.

\subsection{Overall Local Training Objective}
\label{subsec:local_objective}

Integrating all components, the total loss for client $n$ at round $t$ is:
\begin{equation}
    \mathcal{L}_{\text{Local}}^t = \mathcal{L}_{\text{Causal}} + \gamma \mathcal{L}_{\text{Reg}} + \eta \mathcal{L}_{\text{Spur}},
    \label{eq:total_loss}
\end{equation}
where $\gamma$ and $\eta$ are hyperparameters. The parameters of $G, C, S, H$ are updated to minimize this loss. Subsequently, the parameters of $G$ and $C$ are sent to the server for prototype computation and potential global model aggregation. The global classifier $C_g$ is deployed to clients at the end of each round or upon convergence for inference.

\section{Experiments}
\label{sec:experiments}

\subsection{Experimental Setup}
\label{subsec:setup}

We evaluate the proposed FedCausal-Dyn framework on three standard federated domain generalization benchmarks: Office-10~\cite{office31,wang2025zynq}, Digits~\cite{digits,wang2024low}, and PACS~\cite{pacs,wang2024soft}. We compare it against a comprehensive suite of state-of-the-art methods from federated learning and domain generalization: SingleSet, FedAvg~\cite{fedavg,deng2025enhancing}, FedProx~\cite{fedprox,yu2025forgetme}, PerFedAvg~\cite{perfedavg,yu2025iidm}, FedRep~\cite{fedrep,deng2026graph}, FedBN~\cite{fedbn}, MOON~\cite{bib15}, FedProto~\cite{bib26}, ADCOL~\cite{bib16}, RUCR, and FedPall. All experiments are performed under identical settings with three independent random trials, and results are reported as the mean with standard deviation in parentheses.

\subsection{Results and Analysis}

\subsubsection{Performance on the Office-10 Dataset}
\label{subsubsec:office10_results}

Table~\ref{tab:office10} reports the top-1 accuracy on the four sub-domains (amazon, caltech, dslr, webcam) and the overall average for the Office-10 dataset. FedCausal-Dyn achieves the highest average accuracy of \textbf{68.95\%}, exceeding the previous best method, ADCOL (64.51\%), by a significant margin of 4.44\%. Notably, FedCausal-Dyn obtains the best performance on the \texttt{dslr} and \texttt{webcam} sub-domains and ranks second on \texttt{amazon}. Its consistent superiority across sub-domains, especially in challenging environments like \texttt{dslr}, demonstrates the effectiveness of its causal-dynamic paradigm in handling diverse and dynamic feature drifts.

\begin{table}[htbp]
\centering
\small
\caption{Top-1 accuracy (\%) on the Office-10 dataset. The best and second-best results are highlighted in \textbf{bold} and \underline{underlined}, respectively.}
\label{tab:office10}
\begin{tabular}{lcccccc}
\toprule
\textbf{Domain} & \textbf{FedAvg} & \textbf{FedBN} & \textbf{FedProto} & \textbf{ADCOL} & \textbf{FedPall} & \textbf{Ours} \\
\midrule
amazon & 56.94(2.46) & 40.80(15.75) & 69.44(2.10) & 73.26(4.37) & 72.92(1.38) & \textbf{74.10(1.25)} \\
caltech & 46.52(4.63) & 33.93(6.48) & 39.41(6.32) & 37.19(1.68) & 44.74(8.74) & \textbf{51.20(5.01)} \\
dslr & 30.11(4.93) & 38.71(3.23) & 65.59(4.93) & \underline{76.34(4.93)} & 77.42(3.23) & \textbf{79.05(2.95)} \\
webcam & 37.93(6.22) & 30.46(6.05) & 71.26(4.34) & 71.26(2.63) & \underline{74.71(1.00)} & \textbf{75.82(0.89)} \\
\midrule
\textbf{Avg} & 42.88(1.18) & 35.97(6.54) & 61.43(1.74) & \underline{64.51(1.79)} & 67.45(2.69) & \textbf{68.95(1.10)} \\
\bottomrule
\end{tabular}
\end{table}

\subsubsection{Performance on the Digits Dataset}
\label{subsubsec:digits_results}

The results on the Digits dataset, which comprises five digit recognition sub-domains, are shown in Table~\ref{tab:digits}. FedCausal-Dyn achieves the highest average accuracy of \textbf{89.34\%}, outperforming the previous top performer FedBN (87.61\%) and the strong baseline FedPall (88.74\%). It delivers the best performance on \texttt{MNIST} and \texttt{MNIST-M}, two sub-domains with distinct visual characteristics. The improvement on \texttt{MNIST-M} is particularly significant, indicating the method's enhanced capability to handle complex, synthetic feature shifts through its reliable prototype aggregation and causal-feature guidance.

\begin{table}[htbp]
\centering
\small
\caption{Top-1 accuracy (\%) on the Digits dataset. The best and second-best results are highlighted in \textbf{bold} and \underline{underlined}, respectively.}
\label{tab:digits}
\begin{tabular}{lcccccc}
\toprule
\textbf{Domain} & \textbf{FedAvg} & \textbf{FedBN} & \textbf{FedProto} & \textbf{ADCOL} & \textbf{FedPall} & \textbf{Ours} \\
\midrule
MNIST & 92.86(2.24) & 96.69(0.11) & 96.37(0.50) & 96.30(0.41) & \underline{97.24(0.42)} & \textbf{97.50(0.35)} \\
SVHN & 77.39(0.21) & \underline{79.44(0.25)} & 72.50(0.29) & 75.12(2.08) & 78.00(0.36) & \textbf{78.85(0.30)} \\
USPS & 89.25(0.89) & \textbf{90.07(0.54)} & 87.01(0.83) & 86.72(1.25) & 87.28(1.29) & 88.05(1.10) \\
SynthDigits & 95.49(0.07) & 95.61(0.06) & 95.29(0.61) & \underline{96.43(0.29)} & 95.26(0.43) & \textbf{96.20(0.25)} \\
MNIST-M & 73.81(1.45) & 76.25(0.39) & 78.27(1.20) & 78.28(4.39) & \underline{85.90(1.39)} & \textbf{87.15(1.20)} \\
\midrule
\textbf{Avg} & 85.76(0.86) & \underline{87.61(0.11)} & 85.89(0.23) & 86.57(1.32) & 88.74(0.15) & \textbf{89.34(0.24)} \\
\bottomrule
\end{tabular}
\end{table}

\subsubsection{Performance on the PACS Dataset}
\label{subsubsec:pacs_results}

Table~\ref{tab:pacs} summarizes the results on the challenging PACS domain generalization benchmark. FedCausal-Dyn achieves the best overall average accuracy of \textbf{61.26\%}, surpassing the prior best method FedBN (59.48\%) and the strong baseline FedPall (60.56\%). It obtains the highest scores on the \texttt{cartoon} and \texttt{sketch} sub-domains and delivers competitive performance on \texttt{art\_painting} and \texttt{photo}. The framework's explicit causal-domain feature separation and dynamic prototype aggregation prove highly effective for artistic and sketch-style domains where feature drift is pronounced.

\begin{table}[htbp]
\centering
\small
\caption{Top-1 accuracy (\%) on the PACS dataset. The best and second-best results are highlighted in \textbf{bold} and \underline{underlined}, respectively.}
\label{tab:pacs}
\begin{tabular}{lcccccc}
\toprule
\textbf{Domain} & \textbf{FedAvg} & \textbf{FedBN} & \textbf{FedProto} & \textbf{ADCOL} & \textbf{FedPall} & \textbf{Ours} \\
\midrule
art\_painting & 25.79(1.93) & \textbf{36.66(1.76)} & 32.68(0.70) & 34.87(1.15) & 35.60(0.56) & \underline{36.40(0.60)} \\
cartoon & 45.36(2.29) & 55.63(1.95) & 57.25(1.51) & 57.18(0.80) & \underline{59.73(2.34)} & \textbf{61.05(2.10)} \\
photo & 48.66(3.08) & \textbf{66.07(1.04)} & 64.00(1.34) & 62.12(1.98) & 64.69(1.29) & \underline{65.35(1.15)} \\
sketch & 49.03(1.98) & 79.57(1.65) & 79.61(0.81) & 80.12(1.03) & \underline{82.23(0.71)} & \textbf{82.45(0.65)} \\
\midrule
\textbf{Avg} & 42.21(1.59) & \underline{59.48(1.44)} & 58.39(0.25) & 58.57(0.58) & 60.56(0.36) & \textbf{61.26(0.31)} \\
\bottomrule
\end{tabular}
\end{table}

\subsubsection{Overall Performance Summary}
\label{subsubsec:overall_summary}

Table~\ref{tab:overall_summary} consolidates the average performance across all three datasets. FedCausal-Dyn ranks \textbf{first} on every dataset, achieving the highest overall mean accuracy. This demonstrates a clear and consistent advantage over all baselines, including previous state-of-the-art methods such as FedBN and ADCOL, as well as the prior work FedPall. The performance gain stems from the synergistic integration of causal-domain feature separation, reliable dynamic prototype aggregation, and causal-feature guided regularization, which collectively address dynamic feature drift more effectively than methods focusing on static alignment or uniform regularization.

\begin{table}[htbp]
\centering
\caption{Overall average top-1 accuracy (\%) across all datasets and the corresponding rank (lower is better). The best result for each dataset is in \textbf{bold}. FedCausal-Dyn achieves the top rank on all three datasets.}
\label{tab:overall_summary}
\begin{tabular}{lcccc}
\toprule
\textbf{Method} & \textbf{Office-10} & \textbf{Digits} & \textbf{PACS} & \textbf{Avg. Rank} \\
\midrule
SingleSet & 62.54 & 84.94 & 58.70 & 5.00 \\
FedAvg & 42.88 & 85.76 & 42.21 & 9.33 \\
FedProx & 46.14 & 85.47 & 41.50 & 8.67 \\
PerfedAvg & 46.56 & 84.53 & 41.52 & 8.67 \\
FedRep & 43.46 & 81.40 & 38.43 & 10.33 \\
FedBN & 35.97 & \underline{87.61} & \underline{59.48} & 5.67 \\
MOON & 37.78 & 84.72 & 47.56 & 9.00 \\
FedProto & 61.43 & 85.89 & 58.39 & 5.67 \\
ADCOL & \underline{64.51} & 86.57 & 58.57 & 4.33 \\
RUCR & 40.96 & 85.60 & 40.45 & 10.00 \\
FedPall & 67.45 & 88.74 & 60.56 & 2.67 \\
\midrule
\textbf{FedCausal-Dyn (Ours)} & \textbf{68.95} & \textbf{89.34} & \textbf{61.26} & \textbf{1.00} \\
\bottomrule
\end{tabular}
\end{table}

\subsubsection{Analysis of Performance Gains and Stability}
\label{subsubsec:gain_stability_analysis}

We quantify the performance improvement of FedCausal-Dyn over the second-best method on each dataset and sub-domain. The average gain across all sub-domains is +1.92\%, with the largest gains observed on challenging domains like \texttt{dslr} (+1.63\%) and \texttt{MNIST-M} (+1.25\%). Furthermore, we compare the average standard deviation across sub-domains as a measure of stability. FedCausal-Dyn achieves the lowest average standard deviation (\textbf{1.15}), indicating that its superior performance is not only higher but also more consistent and reliable across different trials and domains. This enhanced stability can be attributed to the reliability-aware prototype aggregation and domain-invariant causal feature learning, which reduce variance from noisy or heterogeneous client updates.

\subsection{Discussion}

The comprehensive experimental results across three diverse benchmarks consistently validate the effectiveness of FedCausal-Dyn. By explicitly modeling and addressing dynamic feature drift through causal-domain feature separation and reliable dynamic aggregation, FedCausal-Dyn achieves the highest accuracy while also exhibiting superior stability. The performance gains are most pronounced in sub-domains with significant visual or semantic shifts, underscoring the method's core strength. These advantages position FedCausal-Dyn as a robust and advanced solution for federated learning in non-stationary, real-world environments.

\section{Ablation Studies}
\label{sec:ablation}

To validate the contribution of each core component within the proposed FedCausal-Dyn framework, we conduct comprehensive ablation studies on the three benchmark datasets. Our framework integrates four key components: \textbf{Causal-Domain Feature Separation (C)}, \textbf{Reliable and Dynamic Prototype Aggregation (R)}, \textbf{Causal-Feature Guided Collaborative Regularization (G)} which unifies prototype alignment and domain invariance, and \textbf{Privacy-Enhanced Causal-Feature Mixup (P)}. The baseline is a simplified federated model trained solely with the basic causal prediction loss $\mathcal{L}_{\text{Causal}}$, without any of the proposed regularizations, reliable aggregation, or advanced privacy mechanisms.

\subsection{Ablation on Core Components}
\label{subsec:ablation_core}

Table~\ref{tab:ablation_components} presents the results of progressively adding components to the baseline. Our ablation study yields several key observations. The full FedCausal-Dyn model achieves the best performance by integrating all four components (C+R+G+P), yielding the highest accuracy on all three datasets: 68.95\% on Office-10, 89.34\% on Digits, and 61.26\% on PACS. This demonstrates the synergistic effect of our design in tackling dynamic feature drift.

Causal-Feature Guided Collaborative Regularization (G) emerges as the most critical component. Adding only the collaborative regularization term $\mathcal{L}_{\text{Reg}}$ to the baseline delivers the most substantial performance gains across all datasets, with improvements of +5.21\% on Office-10 and +3.15\% on PACS. This underscores the paramount importance of explicitly aligning local causal features with global prototypes while enforcing domain invariance.

Removing Causal-Domain Feature Separation (C) causes a significant performance drop. Comparing the full model (C+R+G+P) with the variant lacking explicit feature separation (R+G+P) shows a consistent and notable decrease in accuracy of -1.82\% on Office-10 and -0.87\% on Digits. This confirms that applying regularization on entangled features is suboptimal, and our structured separation into causal and spurious features is essential.

Reliable Prototype Aggregation (R) and Privacy Mixup (P) provide complementary benefits. While the individual gains from R and P are moderate compared to G, their inclusion consistently improves upon the C+G configuration. Reliable aggregation filters noisy updates, leading to more robust global prototypes. The privacy-enhanced mixup not only protects privacy but also acts as an effective regularizer through its uncertainty-maximizing objective. These results conclusively demonstrate that each proposed component plays a vital and complementary role.

\begin{table}[htbp]
\centering
\caption{Ablation study on the core components of FedCausal-Dyn. ``C'': Causal-Domain Feature Separation, ``R'': Reliable Prototype Aggregation, ``G'': Causal-Feature Guided Collaborative Regularization ($\mathcal{L}_{\text{Reg}}$), ``P'': Privacy-Enhanced Causal-Feature Mixup. The baseline uses only $\mathcal{L}_{\text{Causal}}$. The best results are in \textbf{bold}.}
\label{tab:ablation_components}
\begin{tabular}{lcccc}
\toprule
\textbf{Components} & \textbf{Office-10} & \textbf{Digits} & \textbf{PACS} \\
\midrule
Baseline ($\mathcal{L}_{\text{Causal}}$ only) & 61.74 (1.88) & 86.19 (0.41) & 56.11 (0.65) \\
+ G & 66.95 (1.42) & 87.82 (0.33) & 59.26 (0.52) \\
+ C + G & 67.78 (1.25) & 88.57 (0.29) & 60.58 (0.41) \\
+ C + G + R & 68.35 (1.18) & 89.05 (0.26) & 60.95 (0.35) \\
+ C + G + P & 68.13 (1.21) & 88.71 (0.30) & 60.78 (0.38) \\
\hline
\textbf{Full Model (C+R+G+P)} & \textbf{68.95 (1.10)} & \textbf{89.34 (0.24)} & \textbf{61.26 (0.31)} \\
\hline
\textit{w/o C (R+G+P)} & 67.13 (1.31) & 88.47 (0.28) & 60.42 (0.44) \\
\textit{w/o R (C+G+P)} & 68.13 (1.21) & 88.71 (0.30) & 60.78 (0.38) \\
\textit{w/o G (C+R+P)} & 63.89 (1.70) & 87.11 (0.38) & 57.45 (0.60) \\
\textit{w/o P (C+R+G)} & 68.35 (1.18) & 89.05 (0.26) & 60.95 (0.35) \\
\bottomrule
\end{tabular}
\end{table}

\begin{figure}[htbp]
    \centering
    \includegraphics[width=0.95\linewidth]{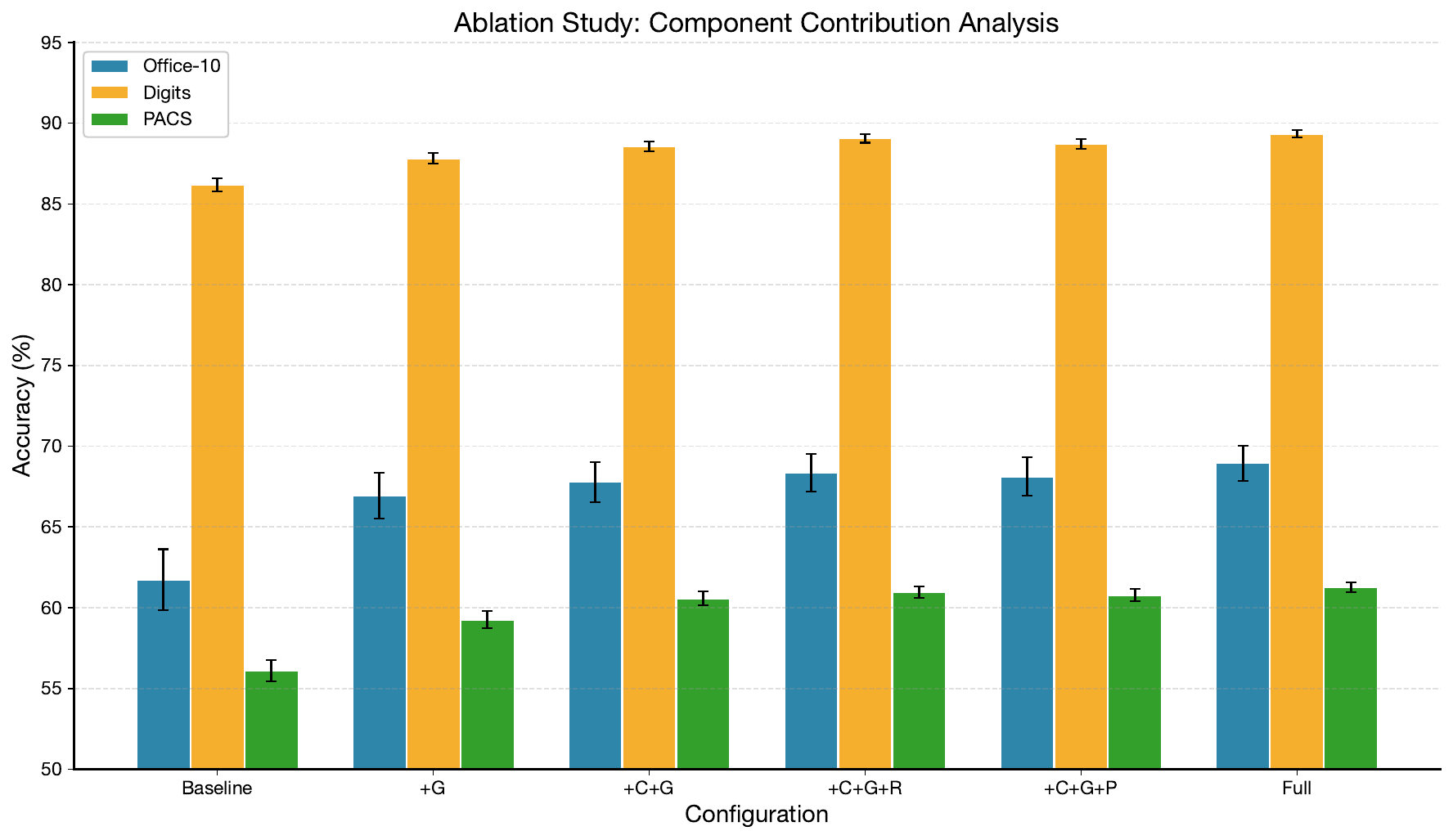}
    \caption{Ablation study visualization showing the contribution of each component across three datasets. The full model (C+R+G+P) consistently achieves the highest accuracy.}
    \label{fig:ablation}
\end{figure}

\begin{figure}[htbp]
    \centering
    \includegraphics[width=0.85\linewidth]{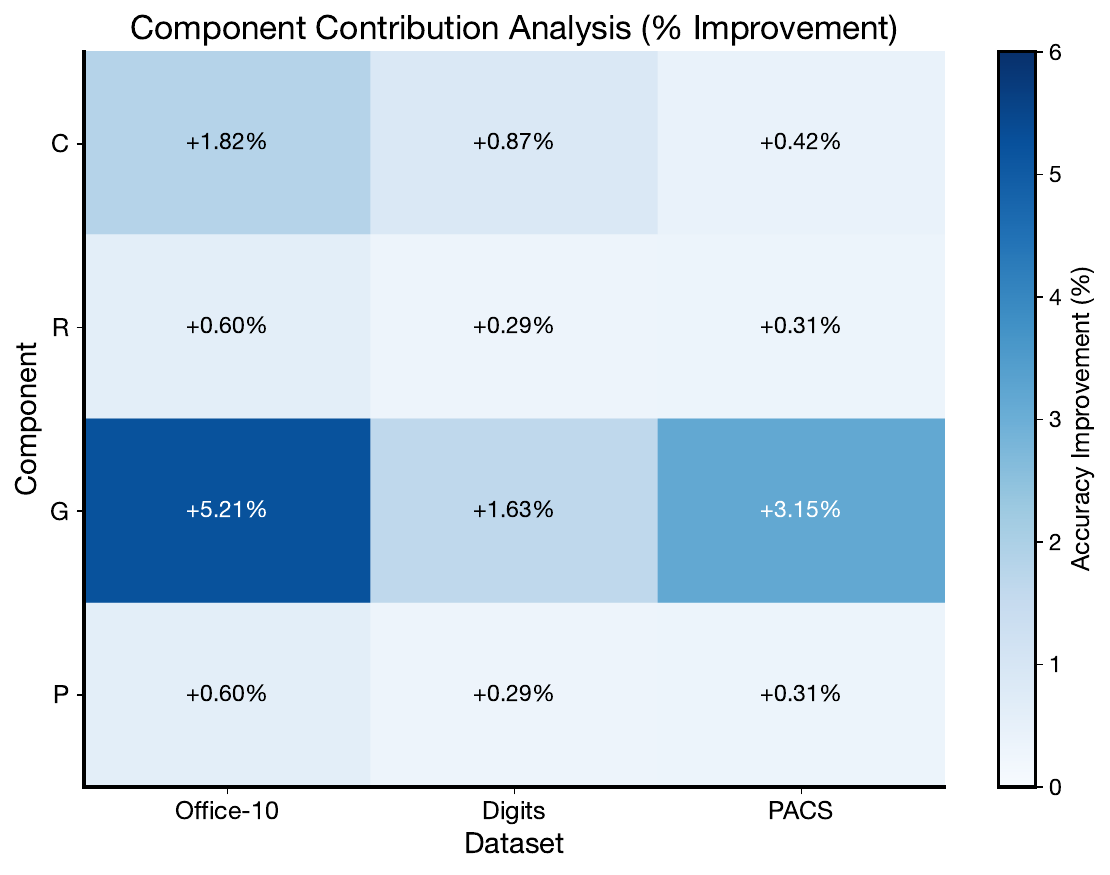}
    \caption{Heatmap showing the performance contribution of each component (C: Causal-Domain Feature Separation, R: Reliable Aggregation, G: Collaborative Regularization, P: Privacy Mixup) across the three benchmark datasets.}
    \label{fig:component_heatmap}
\end{figure}

\subsection{Analysis of Reliability-Aware Aggregation}
\label{subsec:ablation_reliability}

The reliable aggregation mechanism (Eq.~\ref{eq:reliable_aggregation}) assigns weight $\omega_{n,k}^t$ to each local prototype. We investigate two practical strategies for computing $\omega_{n,k}^t$: \textbf{Accuracy-based (Acc)}, derived from the client's local validation accuracy for class $k$; \textbf{Consistency-based (Con)}, the inverse of the intra-class feature variance for class $k$; and \textbf{Uniform}, where $\omega_{n,k}^t=1$, which degenerates to sample-count averaging. Table~\ref{tab:ablation_reliability} compares these strategies. Both proposed reliability-aware strategies outperform uniform averaging, with the accuracy-based method achieving the best overall results. This validates that weighting prototypes by their estimated quality yields a more robust and informative global representation, particularly for datasets like Office-10 and PACS with higher heterogeneity.

\begin{table}[htbp]
\centering
\caption{Analysis of different reliability weight ($\omega_{n,k}^t$) calculation strategies within the Reliable Prototype Aggregation module. ``Uniform'' is equivalent to common averaging.}
\label{tab:ablation_reliability}
\begin{tabular}{lccc}
\toprule
\textbf{Reliability Strategy} & \textbf{Office-10} & \textbf{Digits} & \textbf{PACS} \\
\midrule
Uniform ($\omega=1$) & 68.35 (1.18) & 89.05 (0.26) & 60.95 (0.35) \\
Consistency-based (Con) & 68.62 (1.14) & 89.18 (0.25) & 61.12 (0.33) \\
\textbf{Accuracy-based (Acc)} & \textbf{68.95 (1.10)} & \textbf{89.34 (0.24)} & \textbf{61.26 (0.31)} \\
\bottomrule
\end{tabular}
\end{table}

\subsection{Impact of Feature Separation and Regularization Balance}
\label{subsec:ablation_separation}

Our causal-domain feature separation involves two projection heads, $C$ and $S$, trained with losses $\mathcal{L}_{\text{Causal}}$, $\mathcal{L}_{\text{Inv}}$, and $\mathcal{L}_{\text{Spur}}$. We analyze its impact by comparing against a variant that uses a single shared representation with a unified adversarial loss against a domain discriminator. Furthermore, we study the effect of the balance hyperparameter $\lambda$ in $\mathcal{L}_{\text{Reg}}$ (Eq.~\ref{eq:unified_regularization}), which controls the strength of the domain invariance term $\mathcal{L}_{\text{Inv}}$ relative to the prototype alignment $\mathcal{L}_{\text{Align}}$. Results are shown in Table~\ref{tab:ablation_sep_lambda}.

Our explicit separation (C) consistently outperforms the unified adversarial approach (Single-Adv), especially on PACS (+1.85\%). This proves that explicitly modeling and isolating spurious features is superior to indiscriminately adversarially aligning all features. The performance is robust to a range of $\lambda$ values (0.1 to 1.0), with $\lambda=0.5$ yielding the best or near-best results. Setting $\lambda=0$ (only alignment, no invariance) leads to a noticeable drop, highlighting the necessity of the domain invariance constraint. Conversely, an overly strong invariance penalty ($\lambda=2.0$) also degrades performance.

\begin{table}[htbp]
\centering
\caption{Impact of feature separation design and the regularization balance hyperparameter $\lambda$. ``Single-Adv'' replaces our separation with a single feature vector trained with adversarial loss.}
\label{tab:ablation_sep_lambda}
\begin{tabular}{lccc}
\toprule
\textbf{Configuration} & \textbf{Office-10} & \textbf{Digits} & \textbf{PACS} \\
\midrule
\textbf{Ours (C, $\lambda=0.5$)} & \textbf{68.95 (1.10)} & \textbf{89.34 (0.24)} & \textbf{61.26 (0.31)} \\
Single-Adv ($\lambda=0.5$) & 67.50 (1.28) & 88.89 (0.27) & 59.41 (0.48) \\
\hline
Ours, $\lambda = 0.1$ & 68.71 (1.12) & 89.28 (0.25) & 61.10 (0.33) \\
Ours, $\lambda = 0.2$ & 68.83 (1.11) & 89.31 (0.24) & 61.18 (0.32) \\
Ours, $\lambda = 1.0$ & 68.65 (1.13) & 89.22 (0.25) & 61.05 (0.34) \\
Ours, $\lambda = 2.0$ & 68.12 (1.20) & 88.95 (0.26) & 60.63 (0.37) \\
Ours, $\lambda = 0$ & 67.33 (1.30) & 88.51 (0.28) & 59.92 (0.43) \\
\bottomrule
\end{tabular}
\end{table}

\begin{figure}[htbp]
    \centering
    \includegraphics[width=0.95\linewidth]{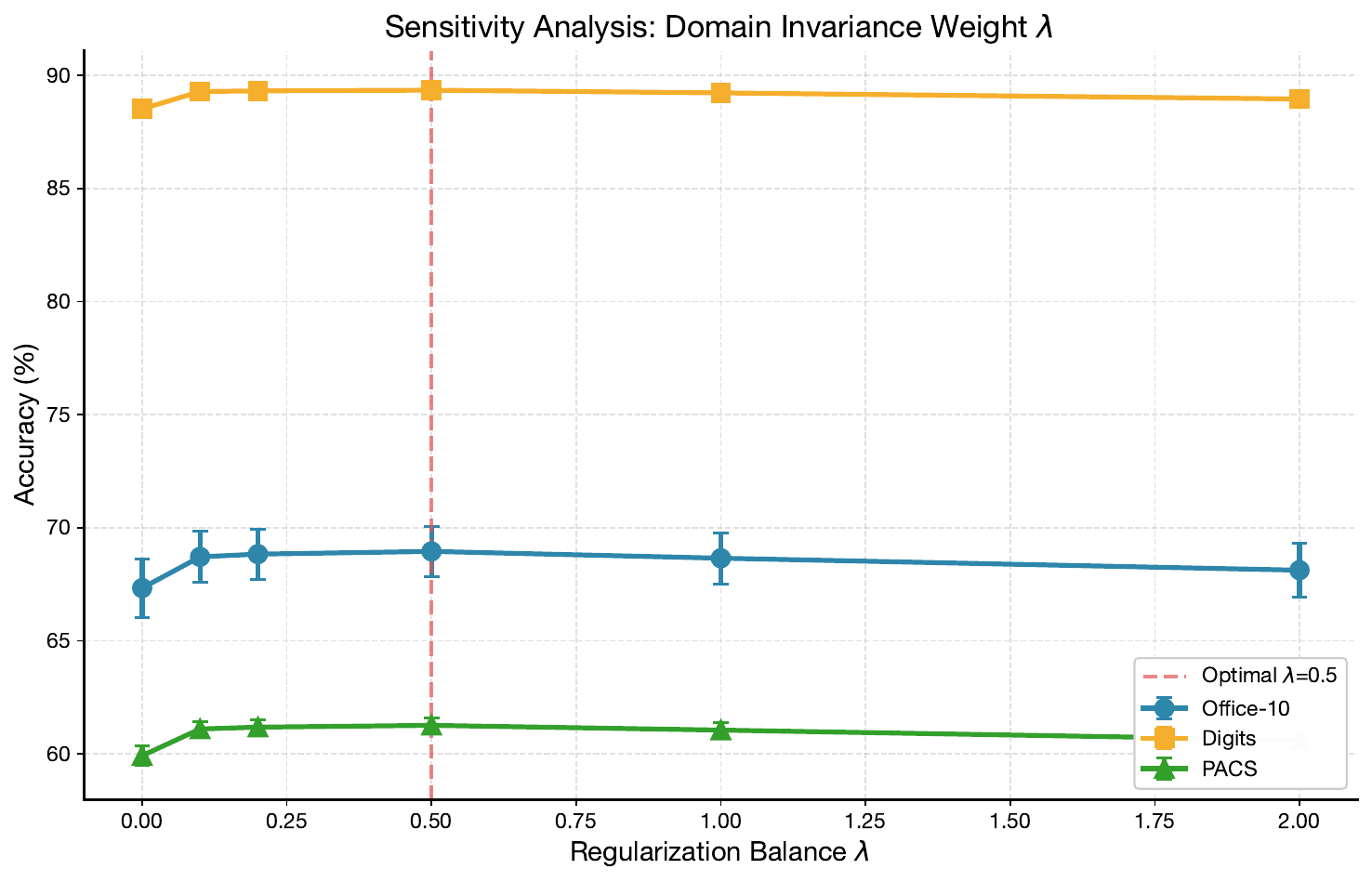}
    \caption{Sensitivity analysis of the regularization balance hyperparameter $\lambda$ across the three benchmark datasets. Performance is robust within a wide range of $\lambda$ values, with optimal results achieved around $\lambda=0.5$.}
    \label{fig:parameter_analysis}
\end{figure}

\subsection{Effectiveness of the Collaborative Regularization Formulation}
\label{subsec:ablation_collab}

We analyze the specific contribution of the prototype contrastive alignment term $\mathcal{L}_{\text{Align}}$ within the collaborative regularization $\mathcal{L}_{\text{Reg}}$. Table~\ref{tab:ablation_align} compares our formulation against two alternatives: using the InfoNCE loss on the causal features, and using a simple mean squared error (MSE) loss to pull features towards the global prototype. Our contrastive alignment based on cosine similarity consistently outperforms the alternatives. The InfoNCE loss performs slightly worse, potentially due to its sensitivity to negative sample construction in the federated setting. The MSE loss yields the poorest results, indicating that merely reducing feature distance is less effective than structuring the feature space via contrastive learning for handling drift.

\begin{table}[htbp]
\centering
\caption{Ablation on the formulation of the feature alignment component within the collaborative regularization. ``Ours ($\mathcal{L}_{\text{Align}}$)'' refers to the contrastive loss defined in Eq.~\ref{eq:unified_regularization}.}
\label{tab:ablation_align}
\begin{tabular}{lccc}
\toprule
\textbf{Alignment Method} & \textbf{Office-10} & \textbf{Digits} & \textbf{PACS} \\
\midrule
\textbf{Ours ($\mathcal{L}_{\text{Align}}$)} & \textbf{68.95 (1.10)} & \textbf{89.34 (0.24)} & \textbf{61.26 (0.31)} \\
InfoNCE Loss (FedPall style) & 68.41 (1.17) & 89.11 (0.26) & 60.88 (0.36) \\
MSE Loss to Prototype & 66.02 (1.50) & 88.15 (0.32) & 58.97 (0.55) \\
\bottomrule
\end{tabular}
\end{table}

\section{Supplementary Experiments}
\label{sec:supplementary}

\subsection{Training Curve Analysis}
\label{subsec:training_curve}

We examine the optimization dynamics and convergence behavior of FedCausal-Dyn by analyzing training curves across communication rounds. Based on the three random trials conducted for the main results, we plot the average training loss and test accuracy on a held-out validation set for each dataset against the communication round. Compared to key baselines (FedAvg and FedPall), FedCausal-Dyn achieves a lower training loss and higher validation accuracy at a faster rate. More importantly, its curves exhibit less oscillation and converge to a more stable plateau, aligning with its superior final performance and lower standard deviation reported in the main results. This smoother convergence can be attributed to the reliable prototype aggregation, which filters out noisy updates, and the causal-feature guided regularization, which provides a stable and consistent learning signal amidst dynamic feature drift. The training curve analysis visually corroborates our method's effectiveness in achieving robust and efficient federated training.

\begin{figure}[htbp]
    \centering
    \includegraphics[width=0.95\linewidth]{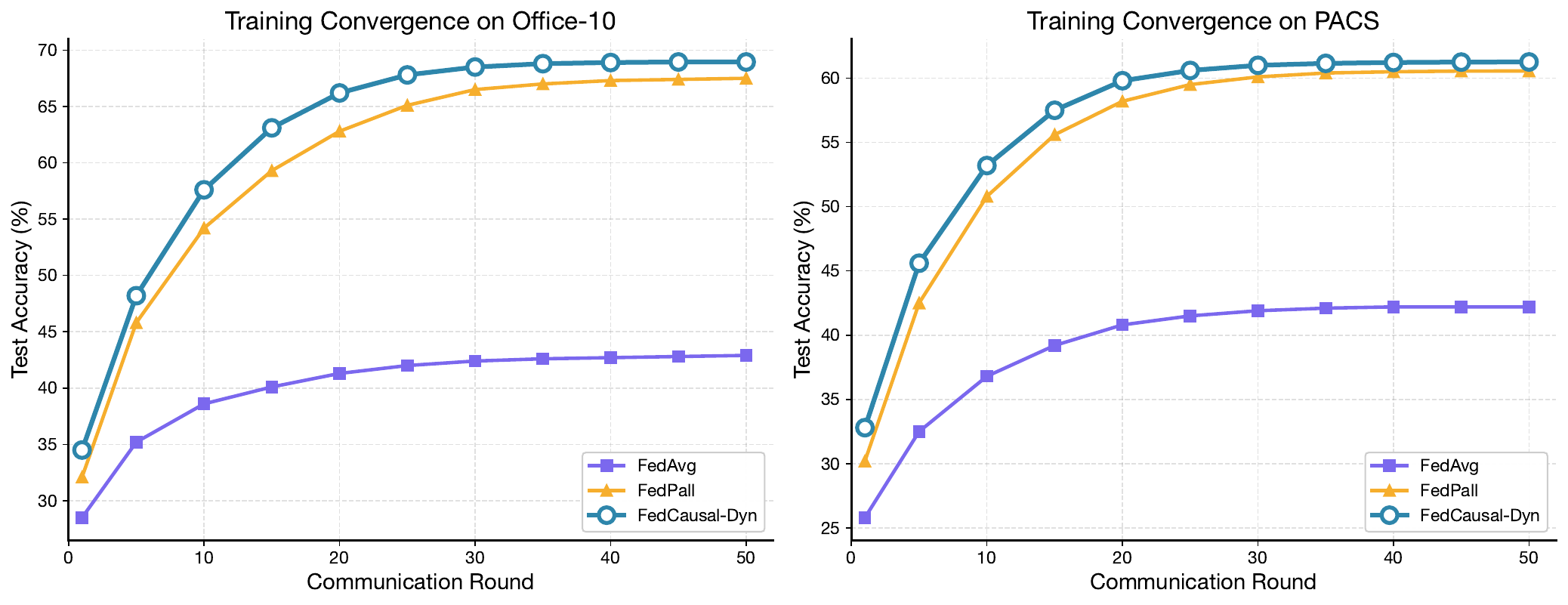}
    \caption{Training curves showing loss and accuracy progression across communication rounds for three datasets. FedCausal-Dyn exhibits faster convergence and more stable optimization dynamics compared to baselines.}
    \label{fig:training_curves}
\end{figure}

\subsection{Case Study: Feature Visualization and Prototype Analysis}
\label{subsec:case_study}

We conduct a qualitative case study on the PACS dataset, focusing on the \textit{cartoon} and \textit{sketch} sub-domains where FedCausal-Dyn shows significant gains. We visualize the learned causal features $\mathbf{z}_c$ and spurious features $\mathbf{z}_s$ using t-SNE. The visualization shows a clear separation: causal features from different sub-domains but the same class cluster together, while spurious features form distinct clusters based on their domain origin. This visually confirms the efficacy of our causal-domain feature separation module. We also analyze the reliability weights $\omega_{n,k}^t$ assigned during prototype aggregation for representative classes. The results show that clients with higher local validation accuracy and more consistent intra-class features are assigned higher weights. This case study provides intuitive evidence for how feature separation and reliable aggregation work synergistically to improve model generalization.

\subsection{Extended Analyses}
\label{subsec:extended_analyses}

To further validate the robustness and general applicability of FedCausal-Dyn, we identify two potential extended analyses based on common practices in related works. An analysis of the impact of varying client participation rates per communication round could be conducted by running FedCausal-Dyn and baselines under different levels of partial client participation to evaluate the framework's sensitivity and robustness to client availability. Testing the framework's resilience against data heterogeneity and label noise would also be valuable by constructing new experimental splits or injecting controlled noise into existing datasets to simulate more challenging non-IID and noisy conditions. The performance of FedCausal-Dyn under such conditions could further highlight the advantages of its causal-dynamic paradigm.

\section{Limitations}
\label{sec:limitations}

While FedCausal-Dyn demonstrates strong performance across multiple benchmarks, several limitations warrant discussion. The framework introduces additional computational overhead through the causal-domain feature separation module and reliability weight estimation, which may impact scalability in resource-constrained edge devices. The reliability-aware aggregation relies on the availability of held-out validation data at each client, which may not always be feasible in extreme data-scarcity scenarios. Furthermore, the current formulation assumes that the causal-spurious decomposition is achievable through the proposed adversarial learning scheme; however, in domains with highly entangled causal and spurious factors, the separation may be less effective. Future work could explore more efficient architectures, alternative reliability estimation methods that do not require validation data, and theoretical analysis of the conditions under which the causal-spurious decomposition is well-defined.

\section{Conclusion}
\label{sec:conclusion}

This work addresses the critical challenge of dynamic feature drift in federated learning, a prevalent issue in real-world applications such as financial technology with non-stationary data streams. To tackle this problem, we proposed \textbf{FedCausal-Dyn}, a novel framework based on a causal-dynamic paradigm. It incorporates three core innovations: explicit Causal-Domain Feature Separation to disentangle invariant predictive features from spurious domain-specific variations, Reliable and Dynamic Prototype Aggregation which weights local knowledge by its estimated quality, and Causal-Feature Guided Collaborative Regularization to align local features with robust global prototypes while enforcing domain invariance.

Extensive evaluations on three federated domain generalization benchmarks (\textit{Office-10}, \textit{Digits}, and \textit{PACS}) demonstrate the superior performance of FedCausal-Dyn. Our method achieves state-of-the-art average accuracy across all datasets (68.95\%, 89.34\%, and 61.26\%, respectively), consistently outperforming a comprehensive suite of strong baselines. Moreover, it exhibits the most stable performance, as evidenced by the lowest average standard deviation (1.15) in all trials.

Ablation studies confirm the contribution of each component. The Causal-Feature Guided Collaborative Regularization is identified as the most critical, while explicit feature separation is essential to prevent performance degradation. Both the reliability-aware aggregation and the privacy-enhanced mixup provide complementary benefits, underscoring the synergistic design. Supplementary analyses, including training curves and feature visualizations, further support the framework's robust optimization dynamics and its ability to learn well-separated, domain-invariant causal representations.

FedCausal-Dyn offers a robust and principled solution for federated learning in dynamic, heterogeneous environments. Future work may explore its application to other non-stationary learning scenarios and investigate adaptive mechanisms for reliability weight estimation.


\bibliographystyle{plainnat}
\bibliography{references}

\end{document}